%% file: Template.tex
\documentclass{article}
\usepackage{spconf,amsmath,graphicx}
\usepackage{subcaption}
\usepackage{comment}
\usepackage{amssymb,mathrsfs}
\usepackage{makecell,diagbox,multirow,array}
\usepackage{tikz,pgfplots}
\usepackage{enumitem}
\usepackage{booktabs,threeparttable}
\usepackage{subfiles}

\newcolumntype{?}{!{\vrule width 4\arrayrulewidth}}
\usepackage{graphicx,calc}
\newlength\myheight
\newlength\mydepth
\settototalheight\myheight{Xygp}
\newcommand{\Lang}[1]{$\mathit{L#1}$}       

\pgfplotsset{compat=1.14}

\usepackage{CJKutf8}

\newenvironment{chinese}
	{\begin{CJK*}{UTF8}{gbsn}}
    {\end{CJK*}}
\title{Training a Code-Switching Language Model with Monolingual Data}
\name{Shun-Po Chuang\thanks{This work is sponsored by Ministry of Science and Technology.}, Tzu-Wei Sung, Hung-yi Lee}
\address{Graduate Institute of Communication Engineering, National Taiwan University\\
\texttt{\{f04942141, b03902042, hungyilee\}@ntu.edu.tw}}

\begin{document}
%
\maketitle
\begin{abstract}
A lack of code-switching data complicates the training of code-switching (CS) language
models.
We propose an approach to train such CS language models
on monolingual data only.  
By constraining and normalizing the output projection matrix in RNN-based language
models, we bring embeddings of different languages closer to each other. 
Numerical and visualization results
show that the proposed approaches remarkably improve the performance of CS
language models trained on monolingual data. 
The proposed approaches are comparable or even better than training
CS language models with artificially generated CS data. 
We additionally use unsupervised bilingual word translation to analyze whether
semantically equivalent words in different languages are mapped together.
\end{abstract}
%
\section{Introduction}
\label{sec:intro}
Code-switching (CS), which occurs when two or more languages are used within a
document or a sentence, is widely observed in multicultural areas.
Related research is characterized by a lack of data;
the application of prior knowledge~\cite{zeng2017improving,6639306} or additional
constraints~\cite{li2013improved,ying2014language} would alleviate this issue. 
Because it is easier to collect monolingual data than CS data,
efficiently utilizing a large amount of monolingual data would be a solution
to the lack of CS data~\cite{hamed2017building}.
Recent work~\cite{gonen2018language} attempts to train a CS language
model using fine-tuning.
Similar work~\cite{garg2017dual} integrates two monolingual language models (LMs) by
introducing a special ``switch''  token in both languages when training the LM,
and further incorporating this within automatic speech recognition (ASR).
Other works synthesize additional CS text using the modeled
distribution from the data~\cite{winata2018learn,yilmaz2018acoustic}.
Generative adversarial neural
networks~\cite{goodfellow2014generative,arjovsky2017wasserstein} 
learn the CS point distribution from CS
text~\cite{chang2018code}. 
In this paper, we propose utilizing constraints to bring word embeddings
of different languages closer together in the same latent space, and to normalize each
word vector to generally improve the CS LM.
Similar constraints are used in end-to-end ASR~\cite{khassanov2019constrained}, 
but have not yet been reported for CS language modeling.
Related prior work~\cite{audhkhasi2017direct,settle2019acoustically}
attempts to initialize the word embedding with unit-normalized vectors in ASR
but does not keep the unit norm during training.
Initial experiments on CS data showed that constraining and
normalizing the output projection matrix helps LMs trained on
monolingual data to better handle CS data.

\section{Code-Switching Language Modeling}
\label{sec:proposed}

In our approach,
we use monolingual data only for training;
CS data is for validation and testing only.

\subsection{RNN-based Language Model}
\label{ssec:rnn}
We adopt a recurrent neural network (RNN) based language
model~\cite{Mikolov2010RecurrentNN}. 
Given a sequence of words $[w_1, w_2, \dots, w_T]$,
we obtain predictions $y_i$ by applying transformation $W$ on RNN hidden
states $h_i$ with softmax computation:
\begin{equation} \label{eq:rnn}
\begin{aligned}
    y_i &= \mathrm{softmax}(W h_i)
\end{aligned}
\end{equation}
where $i = 1, 2, \dots, T$ and $h_0$ is a zero vector. 
Specifically, the output projection matrix is denoted by $W \in \mathbb{R}^{V
\times z}$, where $V$ is the vocabulary size and $z$ is the hidden layer size
of the RNN. Gradient descent is then used to update the parameters with a cross entropy
loss function.

Consider two languages \Lang{1} and \Lang{2} in CS language   
modeling: the output projection matrix $W$ is partitioned into 
$W_1$ and $W_2$, with each row indicating the latent representations of each
word in \Lang{1} and \Lang{2} respectively. 
With careful organization, the output projection matrix $W$ can be written as $\begin{bmatrix}W_1 \\ W_2\end{bmatrix}$.
\subsection{Constraints on Output Projection Matrix}  
\label{ssec:constraint}
\begin{figure}[!t]
    \centering
    \resizebox{\linewidth}{!}{
        \includegraphics[width=0.05\textwidth]{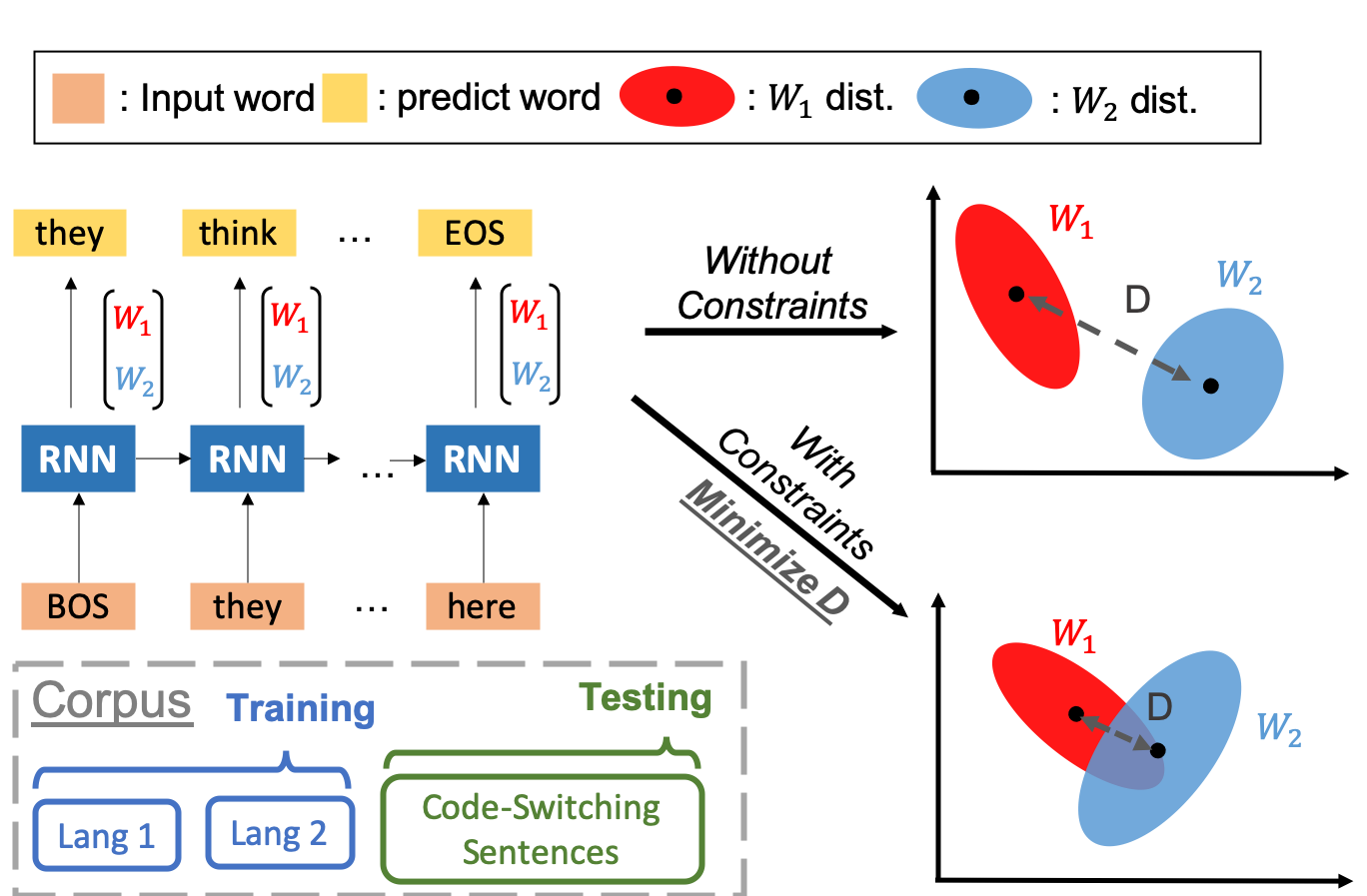}
    }
	 \caption{Overview of proposed approach. Brackets indicate
	 concatenation operation between $W_{1}$ and $W_{2}$}
    \label{fig:method}
\end{figure}

By optimizing the LM with \Lang{1} and \Lang{2} monolingual data,
it is possible to improve the perplexity on both sides.
Word embedding distributions have arbitrary shapes based on their
language characteristics. 
Without seeing bilingual word pairs, however,
the two distributions may converge into their own shape without correlating to each other.
It is difficult to train an LM to switch between languages.
To train an LM with only monolingual data,
we assume that overlapping embeddings benefit CS language modeling.
To this end,
we attempt to bring word embeddings of \Lang{1} and \Lang{2}, that is $W_1$ and
$W_2$, closer to each other.
We constrain $W_1$ and $W_2$ in the two ways;
Fig.~\ref{fig:method} shows an overview of the proposed approach.

\subsubsection{Symmetric Kullback–Leibler Divergence}
\label{ssec:skld}
Kullback–Leibler divergence (KLD) is a well-known measurement of the
distance between two distributions.
Minimizing the KLD between language distributions overlaps
the embedding space semantically.
We assume that both $W_1$ and $W_2$ follow a $z$-dimensional multivariate Gaussian
distribution, that is,
$$
\begin{aligned}
& W_1 \sim N(\mu_1, \Sigma_1), 
& W_2 \sim N(\mu_2, \Sigma_2)
\end{aligned}
$$
where $\mu_1, \mu_2 \in \mathbb{R}^z$ and $\Sigma_1, \Sigma_2 \in \mathbb{R}^{z
\times z}$ are the mean vector and co-variance matrix for $W_1$ and $W_2$
respectively.
Based on the assumption of Gaussian distribution, we can easily compute KLD
between $W_1$ and $W_2$.
Due to the asymmetric characteristic of KLD,
we adopt the symmetric form of KLD (SKLD),
that is,
the sum of KLD between $W_1$ and $W_2$ and that between $W_2$ and
$W_1$:
\begin{align*}
    L_{\mathit{SKLD}} 
             &= \frac{1}{2}\left[tr(\Sigma^{-1}_1\Sigma_2 + \Sigma^{-1}_2\Sigma_1)\right. \\
        &\:\:\:\:\left.+ (\mu_1 - \mu_2)^T (\Sigma^{-1}_1 + \Sigma^{-1}_2) (\mu_1 - \mu_2) - 2z\right].
\end{align*}
\subsubsection{Cosine Distance}
\label{ssec:cd}
Cosine distance (CD) is a common measurement for semantic evaluation.
By minimizing CD,
  we are attempting to bring the semantic latent space of languages closer.
Similar to SKLD, 
we compute the mean vector $\mu_1$ and $\mu_2$ of $W_1$ and $W_2$
respectively, and CD between two mean vectors is obtained as
$$
    L_{\mathit{CD}} = 1 - \frac{\mu_1 \cdot \mu_2}{\|\mu_1\| \|\mu_2\|},
$$
where $\|\cdot\|$ denotes the $\ell^2$ norm.
We hypothesize the latent representation of each word in \Lang{1} and
\Lang{2} is distributed in the same semantic space and overlaps by
minimizing SKLD or CD.
\subsection{Output Projection Matrix Normalization} 
\label{sec:normalizing}
Apart from the constraints from Section~\ref{ssec:constraint}, we propose
normalizing the output projection matrix, that is, each word representation is
divided by its $\ell^2$ norm to possess unit norm. Note that normalization is
independent of constraints, and can be applied together.

In normalization, we consider 
semantically equivalent words $w_j$ and $w_k$: the cosine similarity between
their latent representation $v_j$ and $v_k$ should be 1, implying the angle between 
them is 0, that is, they have the same orientation. By Eq.~(\ref{eq:rnn}), we observe
that the probabilities $y_{i,j}=\frac{\exp(v_j \cdot h_i)}{\sum_{m=1}^V
\exp(v_m \cdot h_i)}$ and $y_{i,k}=\frac{\exp(v_k \cdot h_i)}{\sum_{m=1}^V
\exp(v_m \cdot h_i)}$ are not necessarily equal because the magnitude of $v_j$
and $v_k$ might not be the same. However, being a unit vector, normalization
guarantees that given the same history, the probabilities of two semantically
equivalent words generated by the LM will be equal.
Thus normalization is helpful for clustering semantically
equivalent words in the embedding space, which improves language
modeling in general.

\section{Experimental Setup}
\label{sec:exp_setup}

\subsection{Corpus}
\label{ssec:corpus}
The South East Asia Mandarin-English (SEAME) corpus~\cite{Lyu2010SEAMEAM} was used
for the following experiments.
It can be simply separated into two parts by its literal language.
The first part is monolingual, 
containing pure Mandarin and pure English transcriptions,
the two main languages in this corpus.
The second part is code-switching (CS) sentences, where the transcriptions are a mix of words from the two languages.

The original data consists of \textit{train}, \textit{dev\_man}, and
\textit{dev\_sgn}.\footnote{https://github.com/zengzp0912/SEAME-dev-set}
Each split contains monolingual and CS sentences, 
but \textit{dev\_man} and \textit{dev\_sgn} are dominated by Mandarin and
English respectively.
We held out 1000 Mandarin, 1000 English, and all CS sentences
(because we needed only monolingual data to train the LM) from
\textit{train} as the validation set. 
The remaining monolingual sentences were for the training set.
Similar to prior work~\cite{khassanov2019constrained}, we used
\textit{dev\_man} and \textit{dev\_sgn} for testing, but to balance the 
Mandarin-to-English ratio, we combined them together as the testing set.

\subsection{Pseudo Code-switching Training Data} 
\label{ssec:training_method}
To compare the performance of the constraints and normalization with
an LM trained on CS data, we also introduce
pseudo-CS data training, in which we use monolingual data
to generate artificial CS sentences. Two approaches are used to
generate pseudo-CS data:

\noindent \textbf{Word substitution} Given only monolingual data, we randomly replace a word in monolingual sentences with its corresponding word in the other language based on the substitution
probability to produce CS data.
However, this requires a vocabulary mapping between the two languages.
We thus use the bilingual translated pair mapping provided by 
MUSE~\cite{conneau2017word}.\footnote{https://github.com/facebookresearch/MUSE}
Note that not all translated words are in our vocabulary set.

\noindent \textbf{Sentence concatenation}: 
We randomly sample sentences from different languages from the original corpus and
concatenate them to construct a pseudo-CS sentence which we add to
the original monolingual corpus. 

\subsection{Evaluation Metrics}
\label{ssec:evaluation}
Perplexity (PPL) is a common measurement of language modeling. 
Lower perplexity indicates higher confidence in the predicted target.
To better observe the effects of the techniques proposed above,
we computed five kinds of perplexity on the corpus:
1) \textbf{ZH}: PPL of monolingual Mandarin sentences; 2) \textbf{EN}: PPL
of monolingual English sentences; 3) \textbf{CS-PPL}: PPL of CS sentences; 
4) \textbf{CSP-PPL}: the PPL of CS points, which occur when
the language of the next word is different from current word; 5) \textbf{Overall}:
the PPL of the whole corpus, including monolingual and CS sentences.
Due to the difference between CS-PPL and CSP-PPL,
these perplexities are separately measured.
Clearly, improvements in CS-PPL do not necessarily translate to improvements in CSP-PPL;
as CS sentences often contain a majority of non-CS points, 
CS-PPL is likely to benefit more from improving monolingual perplexity than from 
improving CSP-PPL.

\subfile{tables/results}

\subsection{Implementation}
\label{ssec:implementation}
Due to the limited amount of training data, we adopted only a single recurrent
layer with long short-term memory (LSTM) cells for language 
modeling~\cite{sundermeyer2012lstm}.
The hidden size for both the input projection and the LSTM cells was set to 300. 
We used a dropout of 0.3 for better generalization, and trained the models
using Adam with an initial learning rate of 0.001.
In order to obtain better results, the training procedure was stopped when the
overall perplexity on the validation set did not decrease for 10 epochs. All
reported results are the average of 3 runs.

\section{Results}
\label{sec:results}

\subsection{Language Modeling}
\label{ssec:ppl_results}

The results are in Table~\ref{table:ppl}, which contains
results for
(A) the language model trained with monolingual data only;
(B) word substitution with substitution probability;\footnote{We performed grid
search on the substitution probability and 0.2 achieved the lowest perplexity.}
and
(C) sentence concatenation as mentioned in Section~\ref{ssec:training_method}.
(D), (E), and (F) are the results after applying the normalization from
Section~\ref{sec:normalizing} on (A), (B), and (C) respectively.
Baselines in rows (a)(d)(g) represent the language model trained without
constraints or 
normalization.\footnote{A smoothed 5-gram model was also evaluated, but
it yielded worse performance than the baseline. Due to limited space, we omit
the results here.}
Observing rows (a)(d)(g), 
we observe that learning with pseudo-CS sentences indeed helps considerably
in CS perplexity, which is reasonable because the LM has seen CS cases
during training even though the training data is synthetic.
However, comparing rows (b)(c) with (d) and (g) reveals that
after applying additional constraints,
the LM trained on monolingual data only is comparable or even
better in terms of both monolingual (ZH and EN columns) and CS (CS-PPL and
CSP-PPL columns) perplexity than LMs trained with pseudo-CS data.
Whether using monolingual or pseudo-CS data for training, 
normalizing the output projection matrix generally improves language modeling.
Even trained with monolingual data only,
normalization also improves CSP-PPL, as shown in rows (a) and (j).
Thus we conclude that the monolingual data in our corpus has a similar sentence structure,
and normalization yields a similar latent space, aiding in switching between languages.
After applying SKLD and normalization together,
the CSP-PPL improves, yielding the best results in the monolingual data training case.
The perplexity of CS points is reduced significantly when constraints are applied
on the output projection matrix by minimizing SKLD or CD without
degrading the performance on monolingual data.
Rows (k)(n)(q) also show that combining the SKLD constraint with normalization
results in the best performance on each kind of perplexity over only
monolingual and pseudo-CS data. 

\subfile{figures/visualization}

\subsection{Visualization}
\label{ssec:pca_eval}
In addition to numerical analysis,
we seek to determine if the overlapping level of embedding space is aligned with the
perplexity results.
We applied principal component analysis (PCA) on the output projection matrix,
and then visualized the results on a 2-D plane.
Fig.~\ref{fig:pca} shows the visualized results of different approaches.
Fig.~\ref{sfig:pca:baseline} shows that embeddings of two languages are linear
separable with monolingual data only and without applying any proposed
approach.
After synthesizing pseudo-CS data for training as shown in
Fig.~\ref{sfig:pca:ws},
the embeddings of the two languages are closer than Fig.~\ref{sfig:pca:baseline} but without
excessive overlap.
In
Fig.~\ref{sfig:pca:skld},
they totally overlap.
This corresponds to the numerical results in Table~\ref{table:ppl}:
the closer the embeddings are,
the lower the perplexity is.\footnote{Due to limited space, we do not show the visualization results of
sentence concatenation/CD which is quite similar to 
Fig.~\ref{sfig:pca:ws}/\ref{sfig:pca:skld}.}
\subsection{Unsupervised Bilingual Word Translation} 
\label{ssec:bilingual_word_translation}

To analyze whether words with equivalent semantics in different languages
are mapped together with the proposed approaches, we conducted experiments on
unsupervised bilingual word translation.

Given a word $w$ existing in the same bilingual pair mapping mentioned in
Section~\ref{ssec:training_method}, each word in the other language is ranked
according to the cosine similarity of their embeddings.
If the translated word of $w$ is ranked as the $r$-th candidate, then the
reciprocal rank is $\frac{1}{r}$.
The mean reciprocal rank (MRR) is used as an evaluation metric, which is the
average of the reciprocal ranks; thus the MRR should be less than 1.0, and the closer
to 1.0 the better. The proportion of correct translations
that are in the top 10 candidate list ($r \leq 10$) is also reported as
``P@10''~\cite{Xing2015NormalizedWE}.
In order to mitigate the degradation in performance caused by low-frequency
words, we selected words only with a frequency greater than 80, resulting in 
about 200 vocabulary words in Mandarin and English respectively, and 55 bilingual
pairs used for unsupervised bilingual word translation.

The results of bilingual word translation are in Table~\ref{table:mrr}.
We see performance for Mandarin-English translation (column (A))
in both MRR and P@10 that is worse than that in the reverse direction (column (B)).

Row (i) demonstrates that the unconstrained baseline performs poorly,
whereas additional constraints and normalization in
rows (ii) and (iii) yield significantly improved MRR and P@10 compared
with row (i).
This suggests that constraints and normalization for CS language modeling 
indeed enhance semantic mapping. 

\subfile{tables/mrr}

\subfile{tables/sentences}

\subsection{Sentence Generation}
\label{ssec:setence_gen}
We further evaluated the sentence generation ability of language models
trained only with monolingual data. 
Given part of a sentence, we used the language model to complete the
sentence. 
Two generated sentences and their given inputs are shown in
Table~\ref{table:sentence_generateion}. 
Our best approach with SKLD constraint and normalization, 
listed in column (C), switches languages either from English to Mandarin
(row (i)) or from Mandarin to English (row (ii)). 
However, the baseline model in column (B) fails to code-switch from either
side.

\section{Conclusions}
\label{sec:conclusions}
In this work, we train a code-switching language model with monolingual data by
constraining and normalizing the output projection matrix, yielding improved performance.
We also present an analysis of selected results,
which shows our approaches help monolingual embedding space overlap and improves the measurements on bilingual word translation evaluation.

\bibliographystyle{IEEEbib}
\bibliography{Template}

\end{document}

%% file: tables/results.tex
\begin{table}[t]
\resizebox{\linewidth}{!}{%
\begin{tabular}{l||l|l|l|l|l}
\Xhline{4\arrayrulewidth}
\multicolumn{6}{c}{\textbf{Without normalization}}                             \\ \Xhline{4\arrayrulewidth}
\multicolumn{6}{c}{(A) Monolingual only}                              \\ \hline
                & CS-PPL       & CSP-PPL       & ZH       & EN       & Overall  \\ \hline
(a) Baseline    & 424.80   & 1118.88   & 160.40   & 125.41   & 289.20   \\ \hline
(b) SKLD        & 319.71   & 752.03    & 152.66   & 115.50   & 228.79   \\ \hline
(c )CD          & 328.04   & 778.55    & 150.78   & 112.11   & 231.83   \\ \hline
\multicolumn{6}{c}{(B) Pseudo training data -- Word substitution}      \\ \hline
(d) Baseline    & 348.88   & 884.74    & 156.90   & 119.98   & 246.41   \\ \hline
(e) SKLD        & 298.24   & 671.38    & 157.53   & 120.36   & 219.62   \\ \hline
(f) CD          & 296.84   & 680.19    & 156.09   & 117.10   & 217.56   \\ \hline
\multicolumn{6}{c}{(C) Pseudo training data -- Sentence concatenation} \\ \hline
(g) Baseline    & 340.34   & 831.19    & 160.21   & 138.89   & 248.83   \\ \hline
(h) SKLD        & 289.64   & 628.09    & 152.27   & 126.06   & 216.39   \\ \hline
(i) CD          & 293.98   & 652.35    & 150.76   & 124.05   & 217.83   \\ \Xhline{4\arrayrulewidth}
\multicolumn{6}{c}{\textbf{With normalization}}                                \\ \Xhline{4\arrayrulewidth}
\multicolumn{6}{c}{(D) Monolingual only}                              \\ \hline
                & CS-PPL       & CSP-PPL       & ZH       & EN       & Overall  \\ \hline
(j) Baseline    & 311.77   & 754.21    & 123.28   & 90.71    & 212.44   \\ \hline
(k) SKLD        & 277.94   & 601.58    & 130.11   & 96.27    & 197.15   \\ \hline
(l) CD          & 282.24   & 602.35    & 132.94   & 97.86    & 200.33   \\ \hline
\multicolumn{6}{c}{(E) Pseudo training data -- Word substitution}      \\ \hline
(m) Baseline    & 264.93   & 583.65    & 131.31   & 97.50    & 190.79   \\ \hline
(n) SKLD        & 248.87   & 512.27    & 136.85   & 101.12   & 184.14   \\ \hline
(o) CD          & 251.60   & 517.85    & 138.48   & 101.27   & 185.84   \\ \hline
\multicolumn{6}{c}{(F) Pseudo training data -- Sentence concatenation} \\ \hline
(p) Baseline    & 266.11   & 586.83    & 123.31   & 95.82    & 189.88   \\ \hline
(q) SKLD        & 241.73   & 490.00    & 128.75   & 102.44   & 179.83   \\ \hline
(r) CD          & 247.60   & 499.41    & 128.91   & 103.90   & 183.49   \\ \hline
\end{tabular}%
}
\caption{ZH, EN, CS-PPL, CSP-PPL and overall perplexity on testing set}
\label{table:ppl}
\vspace{-6mm}
\end{table}

%% file: figures/visualization.tex
\begin{figure}[ht]
    \centering
    \begin{subfigure}{0.3\columnwidth}
        \centering
		\includegraphics[width=\textwidth]{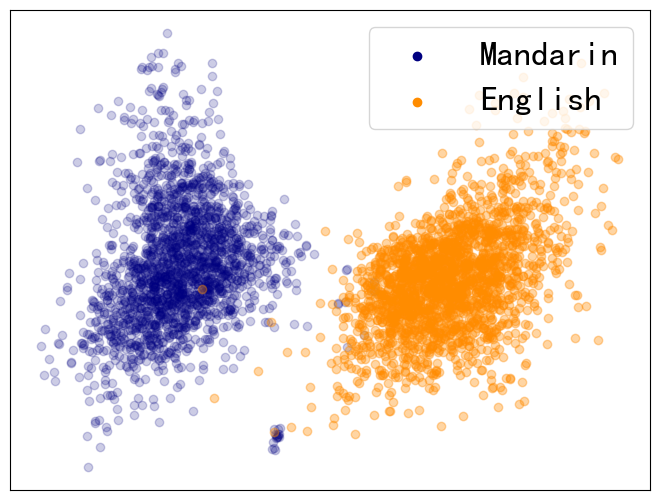}
		\caption{Baseline -- monolingual\\ \centering CSP-PPL: 1102   }  
		\label{sfig:pca:baseline}
    \end{subfigure}
    \begin{subfigure}{0.3\columnwidth}
        \centering
		\includegraphics[width=\textwidth]{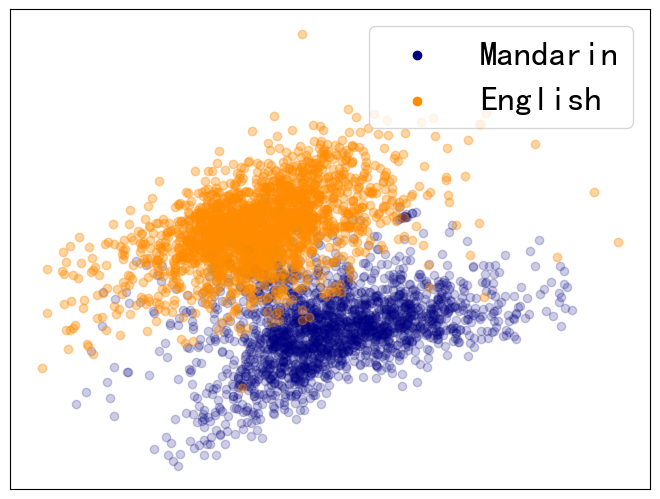}
		\caption{Baseline -- word substitution\\ \centering CSP-PPL: 877    }   
		\label{sfig:pca:ws}
    \end{subfigure}
    \begin{subfigure}{0.3\columnwidth}
        \centering
		\includegraphics[width=\textwidth]{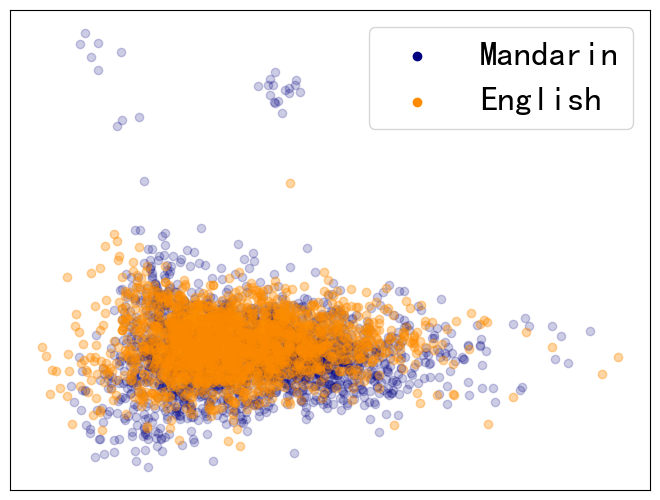}
		\caption{SKLD -- monolingual\\ \centering CSP-PPL: 750   }   
		\label{sfig:pca:skld}
    \end{subfigure}
	 \caption{PCA visualization with different training strategies. Note that
	 the figures are plotted from 1 out of 3 runs.}
    \label{fig:pca}
\end{figure}

%% file: tables/mrr.tex
\begin{table}[t]
\resizebox{\linewidth}{!}{%
\begin{tabular}{l||l|l|l|l}
\toprule
\multirow{2}{*}{\diagbox[width=5.0cm]{Approach}{Metric}} & \multicolumn{2}{l|}{(A) Mandarin $\rightarrow$ English} & \multicolumn{2}{l}{(B) English $\rightarrow$ Mandarin} \\ \cline{2-5} 
                                                         & MRR                     & P@10                  & MRR                     & P@10                  \\ \hline
(i) Baseline                                             & 0.0274                  & 5.4\%                         & 0.0718                  & 20.0\%                        \\ \hline
(ii) + Normalization                                     & 0.0554                  & 14.5\%                        & 0.0885                  & 23.6\%                        \\ \hline
(iii) SKLD + normalization                               & 0.1024                  & 21.8\%                        & 0.1496                  & 30.9\%                        \\ \bottomrule
\end{tabular}%
}
\caption{Results for unsupervised bilingual word translation using different
approaches, all with monolingual training: translation (A) from Mandarin to
English and (B) from English to Mandarin.}
\label{table:mrr}
\end{table}

%% file: tables/sentences.tex
\begin{table}[t]
\resizebox{\columnwidth}{!}{%
\begin{tabular}{l|l||l|l}
\toprule
\multicolumn{2}{l||}{\textbf{(A) Input}}                      & \textbf{(B) Baseline}                                                        & \textbf{(C) SKLD + normalization}                                                                         \\ \hline
(i)  & \begin{chinese}你\:知道\end{chinese} maybe   & \begin{chinese}你\:知道\end{chinese} maybe i think                 & \begin{chinese}你\:知道\end{chinese} maybe \begin{chinese}你\:要\:去\:那边\:的\:时候\:就\:会\end{chinese} \\
     & (you know maybe)                                & (you know maybe i think)                                        & (you know maybe when you go there you will)                                                  \\ \hline
(ii) & they think \begin{chinese}这里\end{chinese} & they think \begin{chinese}这里\:的\:时候\:我\:就\:会\:去\:了\end{chinese} & they think \begin{chinese}这里\end{chinese} is like a lot of people                            \\
     & (they think here)                         & (when they think here i will go)                                & (they think here is like a lot of people)                                                    \\ \bottomrule
\end{tabular}%
}
\caption{Example generated sentences for different approaches, all with
monolingual training: the CS point is (i) from English to Mandarin,
and (ii) from Mandarin to English. English translation in parentheses.}
\label{table:sentence_generateion}
\end{table}